\documentclass[conference]{IEEEtran}

\usepackage{cite}
\usepackage{amsmath,amssymb,amsfonts,amsthm,mathtools}
\usepackage{booktabs}
\usepackage{multirow}
\usepackage{graphicx}
\usepackage{textcomp}
\usepackage{xcolor}
\usepackage{microtype}
\usepackage[colorlinks=true,linkcolor=blue,citecolor=blue,urlcolor=blue]{hyperref}
\graphicspath{{figures/}}

\newcommand{\dvsp}{DVSP}

\newtheorem{proposition}{Proposition}

\title{Differentiable Virtual-Slot Partitioning for\\
Adaptive Tokenization in Vision Transformers}

\author{\IEEEauthorblockN{Gaoxiang He, Bin Zhu, Bo Xie, Xiaotong Yu, Yi Chen, and Hongfei Yan}
\IEEEauthorblockA{National University of Defense Technology}}

\begin{document}

\maketitle

\begin{abstract}
Adaptive tokenization aims to reduce the compute spent on uninformative image
regions while preserving accuracy. Existing work relies on (i) fixed or frozen
scoring rules, which cannot be trained toward the task loss, or (ii) discrete
masks trained with straight-through estimators, which yield noisy gradients,
while fully differentiable alternatives instead leave the pre-trained regular
grid. We revisit adaptive tokenization from a regular-grid perspective and ask:
can a two-scale quadtree be learned end-to-end without deleting a single token
during training? We show that one coarse token and its four fine children can
be made the two endpoints of a single continuous interpolation. We introduce
Differentiable Virtual-Slot Partitioning (\dvsp{}), which keeps four
\emph{virtual slots} per $2{\times}2$ region and blends their token features
and attention mass with a single weight $\alpha$, exactly reproducing the
discrete grid at $\alpha \in \{0,1\}$. On COCO detection with EVA-02-B,
\dvsp{} retains $97.9\%$ of dense Box AP while removing $64.2\%$ of visual
tokens, giving a $1.61\times$ end-to-end speedup, and it outperforms
entropy-based APT, DART, and uniform lower-resolution dense adaptation at
matched budgets; these gains depend on the conserved attention mass, whose
removal costs $1.59$ Box AP. On a frozen Segmenter it also improves Cityscapes
mIoU, and on frozen DeiT backbones it exceeds entropy selection at identical
per-image token counts.
\end{abstract}


\section{Introduction}
\label{sec:intro}

\begin{figure}[!t]
\centering
\includegraphics[width=\columnwidth]{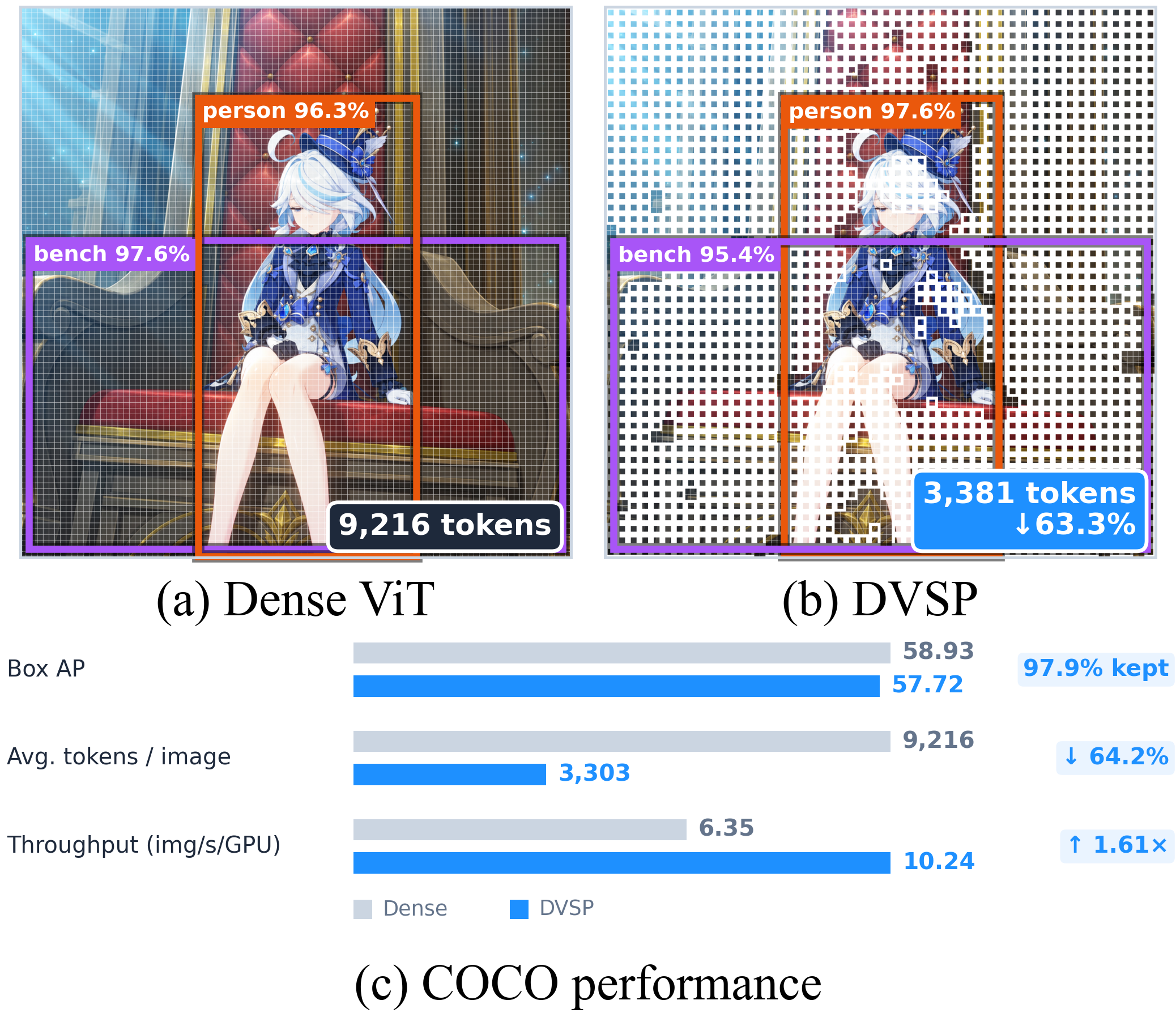}
\caption{\textbf{\dvsp{} overview.} (a) A dense ViT detector processes
$9{,}216$ uniform $16{\times}16$ tokens. (b) \dvsp{} chooses
$16{\times}16$ or $32{\times}32$ tokens per $2{\times}2$ parent with
learned continuous weights, keeping fine tokens on the person and bench.
(c) On COCO val2017 (EVA-02-B), \dvsp{} retains $97.9\%$ Box AP
($58.93\!\to\!57.72$) with $3{,}303$ tokens per image
($\downarrow64.2\%$) and a measured $1.61\times$ speedup.}
\label{fig:teaser}
\end{figure}

Processing every part of an image with a uniform $16{\times}16$ grid is
wasteful. Vision Transformers~\cite{dosovitskiy2021image} spend as much compute
on empty sky and flat roads as they do on small pedestrians. Because attention
and MLP costs grow with the number of tokens, a simple remedy is to mix token
sizes on a regular $2{\times}2$ quadtree grid (Fig.~\ref{fig:teaser}): we can
summarize a flat $2{\times}2$ background region with a single coarse
($32{\times}32$) token while keeping four fine ($16{\times}16$) tokens for
detailed objects. Staying on this regular grid has clear practical advantages:
every token sits on the exact grid the backbone was pre-trained on, the
all-fine state reproduces the original dense model up to floating-point
round-off, and detection heads
can restore a full $H{\times}W$ feature map by simple copying---preserving exact
2D lattice neighborhoods for downstream convolutions---rather than blurry
irregular interpolation.

\begin{figure*}[t]
\centering
\includegraphics[width=\textwidth]{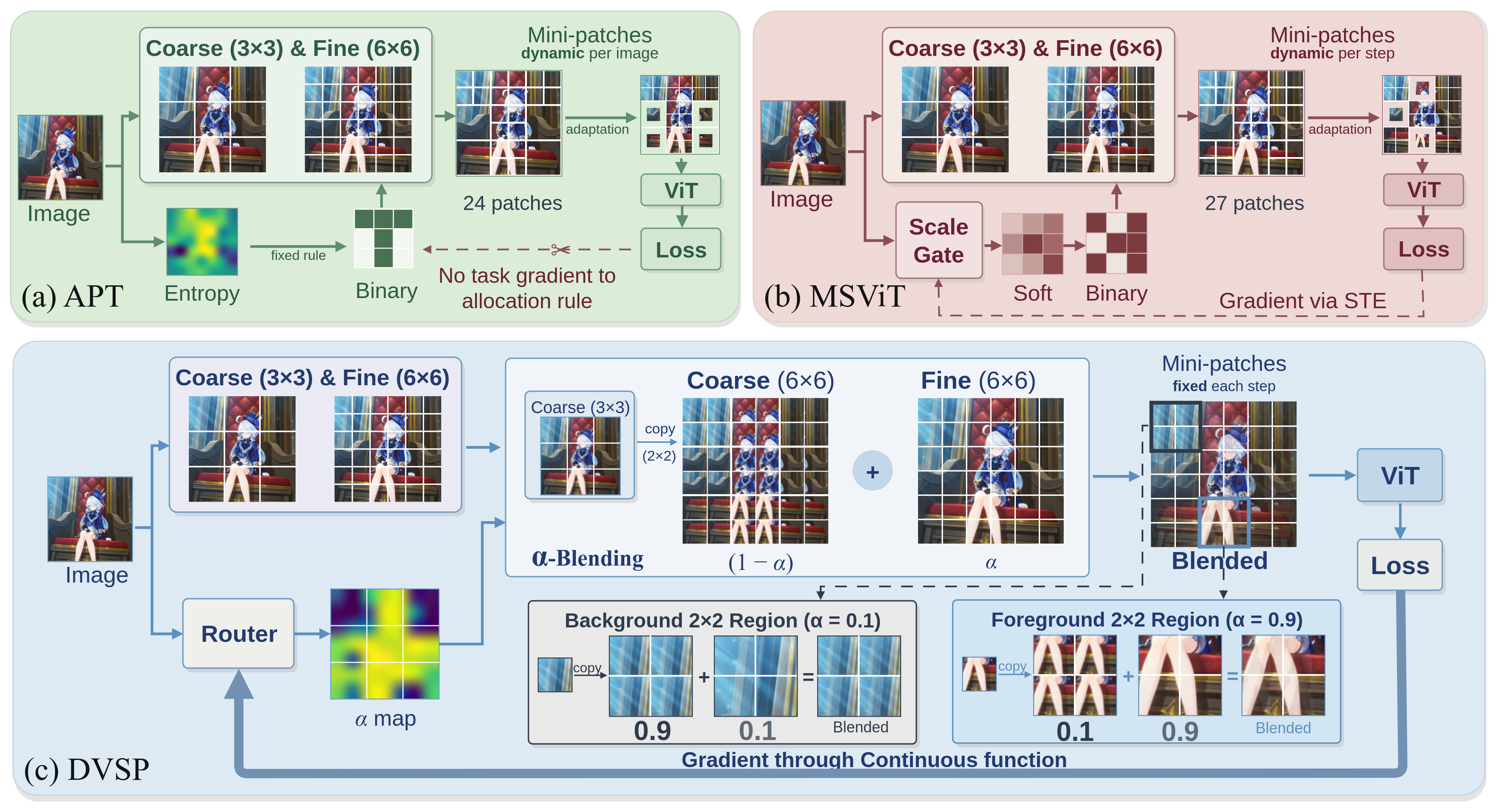}
\caption{\textbf{Token allocation paradigms.} (a) APT picks coarse/fine regions with a fixed image-entropy rule, so task gradients cannot train the router. (b) Discrete-gate methods (e.g., MSViT) gate in the forward pass and rely on straight-through gradients. (c) \dvsp{} keeps four continuous \emph{virtual slots} per parent: one router weight $\alpha$ blends token features and attention mass, so exact task gradients reach the router.}
\label{fig:paradigms}
\end{figure*}

So why hasn't end-to-end quadtree tokenization become standard practice? The
obstacle is learning the discrete $1\text{-vs-}4$ split. As shown in
Fig.~\ref{fig:paradigms}, existing methods have been forced into an awkward
compromise. One camp keeps the regular quadtree grid, but cannot pass exact task
gradients to the router: early works froze the scorer~\cite{ronen2023quadformer}
or trained binary keep/drop masks with noisy straight-through estimators (STE)
and Gumbel relaxations~\cite{rao2021dynamicvit,havtorn2023msvit}, while the
strongest recent quadtree detector, APT~\cite{choudhury2026apt}, gave up on
learning altogether and used a fixed image-entropy rule. The other camp gets
exact analytical gradients only by leaving the pre-trained lattice, allowing
token borders to drift to continuous positions via quantile
cuts~\cite{yin2025dart} or soft superpixels~\cite{aasan2025dhvt}---a concession
that warps pre-trained patch geometry and forces dense prediction heads to
rely on lossy, non-grid resampling.

This clear distinction has led to an implicit agreement that maintaining a regular
grid and achieving native differentiability are inherently conflicting. We argue
that this incompatibility is an illusion: \textbf{the main contributor is not the grid
geometry itself, but the habit of formulating token selection as discrete
masking}. The moment we write a $1\text{-vs-}4$ choice as a binary gate that
deletes three tokens and keeps one, the forward pass undergoes a hard jump in
sequence length. With those three tokens gone from the computational graph, the
exact gradient is lost.

What if a coarse region never deletes three tokens during training? Suppose
instead that every $2{\times}2$ parent region \emph{always} keeps four
\emph{virtual slots} in the training sequence, controlled by a single continuous
weight $\alpha_i \in \mathbb{R}$ (Fig.~\ref{fig:paradigms}(c)). At $\alpha_i{=}1$,
the four slots hold the four fine tokens. As $\alpha_i$ slides toward $0$, the four
slots stay in the sequence, but their token features continuously contract toward the
parent's coarse token.

At first glance, this fixed-slot idea seems to solve the gradient problem---until
we look closely at what happens inside self-attention at $\alpha_i{=}0$. When
all four slots collapse into identical copies of the coarse token, every query
in the image now attends to \emph{four} identical key-value pairs in that parent
region. As a result, a collapsed coarse region exerts \emph{four times} the
attention pull of a single real coarse token. Left unaddressed, the fixed-length
training network does not even compute the same function as the variable-length
test network when $\alpha_i$ is strictly $0$ or $1$.

How can four identical copies in training act exactly like a single token at
test time? The answer is simple: if a coarse parent is split across four slots
at $\alpha_i{=}0$, each slot should carry only \emph{one quarter} of a token's
attention weight. In Differentiable Virtual-Slot Partitioning (\dvsp{}), each
virtual slot interpolates not only \emph{what it says} (\emph{token features}), but also
\emph{how much it weighs} (\emph{attention mass}). By assigning
each slot a mass of $m(\alpha_i) = (1+3\alpha_i)/4$---added simply as a
$+\log m(\alpha_i)$ bias to the attention score---four quarter-mass slots at
$\alpha_i{=}0$ sum up to one unit-mass coarse token, while $\alpha_i{=}1$
recovers four unit-mass fine tokens. This linear mass rule continuously extends
the proportional-attention principle from discrete token
merging~\cite{bolya2023tome}, making our fixed-length $4R$-slot network agree
layer by layer with the real variable-length network at every binary vertex
$\alpha \in \{0,1\}^R$.

Because the training and test networks meet at $0$ and $1$, both training and
deployment become simple. Starting from $\alpha{\equiv}1$, which reproduces the
pre-trained dense ViT up to round-off, we jointly optimize the router
and backbone under the task loss and an $\ell_1$ penalty on $\alpha$. At test
time, we project each region onto its nearest binary vertex, snap the coarse
ones into a single real token, and pack the surviving variable-length sequence
into FlashAttention.

Empirically, \dvsp{} cuts $64.2\%$ of visual tokens on COCO object detection
with EVA-02-B while keeping $97.9\%$ of dense Box AP ($58.93\!\to\!57.72$),
achieving a $1.61\times$ wall-clock speedup and outperforming both heuristic
quadtrees (APT~\cite{choudhury2026apt}, matching its accuracy with
$\sim$3{,}300 vs.\ $\sim$4{,}200 tokens) and quantile-cut tokenizers
(DART~\cite{yin2025dart}).

Our main contributions are:
\begin{itemize}
\item \textbf{Virtual slots with attention mass conservation.} We replace
discrete masking with continuous virtual slots and derive the linear mass rule
$m(\alpha)=(1+3\alpha)/4$, so fixed-length training and native variable-length
execution agree layer by layer at every binary vertex.
\item \textbf{Improved adaptive detection on regular grids.} \dvsp{} achieves
wall-clock speedups on COCO detection without sacrificing the pre-trained
priors or exact lattice geometry of the dense backbone.
\end{itemize}

\section{Related Work}
\label{sec:related}

\paragraph{Heuristic and frozen-scorer quadtrees}
To stay on the pre-trained grid without confronting discrete optimization,
several methods decouple the router from the task loss.
QuadFormer~\cite{ronen2023quadformer} scores regions using frozen lightweight
features, while APT~\cite{choudhury2026apt} thresholds local grayscale image
entropy to assign $32{\times}32$ vs.\ $16{\times}16$ patches. While APT runs
fast via packed variable-length sequences, its accuracy is capped by a basic
mismatch: pixel entropy measures texture contrast, not semantic importance.
Whenever a scene has textured background clutter (like rough walls or foliage)
alongside smooth foreground objects, entropy wastes fine tokens on the
background.

\paragraph{Mask-based discrete gating}
A second group of methods stays on the regular grid and trains a router using
binary keep/drop masks. DynamicViT~\cite{rao2021dynamicvit},
A-ViT~\cite{yin2022avit}, DVT~\cite{wang2021dvt}, and
MSViT~\cite{havtorn2023msvit} make hard binary decisions in the forward pass
and approximate gradients with STE or Gumbel-Softmax in the backward pass.
Because a hard gate cuts the computational graph, the router never receives the
true partial derivative of self-attention with respect to coarse-fine mixing,
producing noisy gradient estimates that struggle to rank region importance.

\paragraph{Off-lattice differentiable tokenization}
To bypass discrete routing decisions, recent differentiable tokenizers relax the
pre-trained $16{\times}16$ lattice, shifting token boundaries to continuous
positions via quantile cuts~\cite{yin2025dart} or soft superpixel
clustering~\cite{aasan2025dhvt}.
While this unlocks analytical spatial gradients, forsaking the regular lattice
imposes a fundamental ceiling on pre-trained transfer and dense prediction.
First, resampling stretched or irregular regions into fixed-size tokens
introduces anisotropic geometric distortion, and their positional encodings must
be interpolated at floating centers, breaking bit-for-bit alignment with the
isotropic $16{\times}16$ patch priors of pre-trained foundation models. Second,
in dense prediction, semantic tokens can be heavily merged only if their spatial
layout can be faithfully restored at the decoder. On a regular quadtree grid,
unmerging via simple index copying preserves the exact 2D lattice topology,
allowing standard $3{\times}3$ convolutions and boundary padding in detection or
segmentation heads to immediately reconstruct fine-grained
localization~\cite{islam2020position,chu2023conditional,xie2021segformer}. In contrast,
projecting irregular tokens back to a dense $H{\times}W$ canvas requires
continuous interpolation, acting as a low-pass filter that degrades
high-frequency object boundaries. \dvsp{} shows that leaving the regular grid is
unnecessary: virtual slots with attention mass provide exact gradients directly
on the standard quadtree grid.

\paragraph{Test-time token merging}
Token Merging (ToMe)~\cite{bolya2023tome} merges similar tokens layer-by-layer
at test time using bipartite matching and proportional attention, where a merged
token representing $s$ original patches receives a $+\log s$ key bias. Because
ToMe follows a fixed, task-agnostic merge schedule without end-to-end training,
heavy merging hurts dense detection accuracy. Conceptually, our attention mass
rule generalizes ToMe's discrete $+\log s$ multiplicity fix into a continuous,
learnable weight over fixed virtual slots.

\section{Method}
\label{sec:method}

The design driver of \dvsp{} is \textbf{how the model executes at inference},
not how it trains. If test time also kept all $4R$ slots, no wall-clock speedup
would occur and no attention re-weighting would ever be needed. Real speedup
requires physically collapsing each coarse $2{\times}2$ region into a single
token and executing a short, variable-length sequence
(Fig.~\ref{fig:paradigms}(c)). We therefore develop \dvsp{} backward from that
target: Sec.~\ref{sec:setup} defines the compact test-time pipeline and lattice
unmerging, Sec.~\ref{sec:slots} constructs the continuous $4R$-slot relaxation
that trains against it, Sec.~\ref{sec:mass} shows why the $4{\to}1$ collapse
necessitates attention mass conservation, and Sec.~\ref{sec:training} gives the optimization and projection rules.

\subsection{Target Variable-Length Inference on a Quadtree Grid}
\label{sec:setup}

A standard Vision Transformer with patch size $16$ maps an image
$x \in \mathbb{R}^{3 \times 16H \times 16W}$ to an $H{\times}W$ grid of fine
tokens ($14{\times}14{=}196$ at $224^2$, or $96{\times}96{=}9{,}216$ at
$1536^2$). Grouping this grid into non-overlapping $2{\times}2$ blocks defines
$R{=}(H/2)(W/2)$ \emph{parent regions}, each covering $J{=}4$ fine \emph{child
patches} $j \in \{1,2,3,4\}$. Each parent $i \in \{1,\dots,R\}$ admits tokens
at two native scales:
\begin{enumerate}
\item \textbf{Fine tokens ($16{\times}16$):} Four standard patch embeddings
$\{f_{i,j}\}_{j=1}^4 \subset \mathbb{R}^d$ at child positions
$\{p_{i,j}^f\}_{j=1}^4$.
\item \textbf{Coarse token ($32{\times}32$):} A single parent token
$c_i \in \mathbb{R}^d$ at parent position $p_i^c$. Following
APT~\cite{choudhury2026apt}, $c_i$ sums the patch embedding of the
half-resolution image and a $2{\times}2$ strided-convolution residual over
$\{f_{i,j}\}_{j=1}^4$ whose final linear layer is zero-initialized:
\begin{equation}
c_i \;=\; \mathcal{E}_{\mathrm{patch}}\!\bigl(\mathrm{Down}_{0.5}(x)_i\bigr)
\;+\; \mathcal{R}_{\theta}\!\bigl(\{f_{i,j}\}_{j=1}^4\bigr).
\nonumber
\end{equation}
\end{enumerate}

\paragraph{Test-time execution pipeline}
At inference time, a lightweight router $g_\phi(x) \in \mathbb{R}^R$ assigns
each parent $i$ a binary switch $b_i \in \{0,1\}$. In our detection
instantiation $g_\phi$ is a small convolutional network that shares no
parameters with the detector: frozen stride-16 MobileNetV3-Large features pass
through $1{\times}1$ fusion stages and a compact head that emits one
unconstrained coordinate per parent (Appendix~\ref{app:details}), and its
$9.1$\,ms routing cost appears explicitly in the end-to-end breakdown of
Fig.~\ref{fig:runtime}. The packed sequence is built in two steps:
\begin{itemize}
\item \textbf{Token packing:} Fine parents ($b_i{=}1$) emit their $4$ fine
tokens $\{(f_{i,j}, p_{i,j}^f)\}_{j=1}^4$, whereas coarse parents ($b_i{=}0$)
physically collapse the $2{\times}2$ block into \emph{one} coarse token
$(c_i, p_i^c)$. This packs the image into a compact variable-length sequence of
\begin{equation}
N(b) \;=\; R \;+\; (J-1)\,\textstyle\sum_{i=1}^R b_i \;=\; R \;+\; 3\textstyle\sum_{i=1}^R b_i
\label{eq:count}
\end{equation}
unit-weight tokens, which executes directly through standard variable-length
FlashAttention across both global and windowed ViT blocks.
\item \textbf{Exact lattice restoration ($\mathcal{U}$):} For dense prediction
heads (detection and segmentation), after the final ViT block a parameter-free
unmerging operator $\mathcal{U}$ restores the full $H{\times}W\times d$ feature
map on the regular grid: if $b_i{=}1$, the $4$ fine output tokens are placed at
their child cells $(i,j)$; if $b_i{=}0$, the single coarse output token is
copied across its $2{\times}2$ child cells, preserving exact 2D lattice
neighborhoods for downstream $3{\times}3$ convolutions.
\end{itemize}

\subsection{Virtual-Slot Interpolation for Differentiable Training}
\label{sec:slots}

While the packed $N(b)$-token graph makes inference fast, training $g_\phi$
directly on it is blocked by the discrete $1\text{-vs-}4$ sequence-length jump
in Eq.~\eqref{eq:count}, which severs $\partial \mathcal{L}_{\mathrm{task}} /
\partial \alpha_i$. To make routing differentiable without leaving the regular
grid, \dvsp{} keeps $J{=}4$ \emph{virtual slots} per parent during training so
the training sequence always has fixed length $4R = H{\times}W$. An
unconstrained router weight $\alpha_i \in \mathbb{R}$
linearly interpolates slot content and position across $[0,1]$:
\begin{align}
\textbf{Content:}\quad
e_{i,j}(\alpha_i) &\;=\; (1-\alpha_i)\,c_i + \alpha_i\, f_{i,j},
\label{eq:content}\\
\textbf{Position:}\quad
p_{i,j}(\alpha_i) &\;=\; (1-\alpha_i)\,p_i^{c} + \alpha_i\, p_{i,j}^{f},
\label{eq:position}
\end{align}
where Eq.~\eqref{eq:position} interpolates additive positional embeddings (or
rotary angles on RoPE backbones). At $\alpha_i{=}1$, the $4$ slots hold the
native fine tokens $\{f_{i,j}, p_{i,j}^f\}_{j=1}^4$; at $\alpha_i{=}0$, all $4$
slots contract to identical copies of the coarse parent $(c_i, p_i^c)$.

\begin{figure}[t]
\centering
\includegraphics[width=\columnwidth]{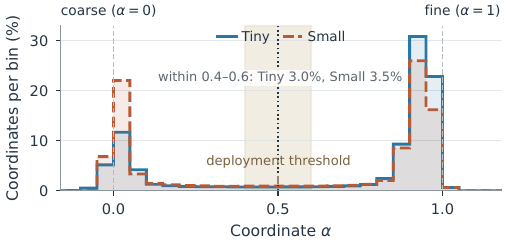}
\caption{\textbf{Learned coordinates after training} on frozen DeiT-Tiny/Small
(ImageNet-1k). Coordinates concentrate at the two binary endpoints; at most
$3.5\%$ remain within $0.4$--$0.6$ of the deployment threshold, and excursions
outside $[0,1]$ are shallow ($6.4\%$/$7.6\%$, almost all within $0.1$ of $0$).}
\label{fig:coords}
\end{figure}

\subsection{Why Test-Time Token Collapse Requires Mass Conservation}
\label{sec:mass}

If inference also ran on all $4R$ virtual slots, Eqs.~\eqref{eq:content}
and~\eqref{eq:position} alone would suffice. However, because test-time
inference physically collapses each coarse parent ($\alpha_i{=}0$) from $4$
virtual slots into \emph{a single physical token} ($4{\to}1$), standard
self-attention breaks training--inference equivalence at $\alpha_i{=}0$: in the
$4R$-slot training sequence, a collapsed parent still occupies $4$ identical
key-value slots $(k_i, v_i)$, so any query $q$ accumulates
$\sum_{j=1}^4 \exp(q^\top k_i / \sqrt{d_h}) v_i = 4\exp(q^\top k_i / \sqrt{d_h}) v_i$
---exerting $4\times$ the attention pull of the single coarse token executed at
test time.

To make the $4$ training slots at $\alpha_i{=}0$ act exactly like $1$ physical
coarse token at test time, we equip each virtual slot $(i,j)$ with a continuous
\emph{attention mass} that interpolates linearly between $m(0){=}1/4$ (four
copies sharing a single unit of coarse mass) and $m(1){=}1$ (four unit-mass fine
tokens):
\begin{equation}
\textbf{Mass:} \qquad
m_{i,j}(\alpha_i) \;=\; \frac{1+3\,\alpha_i}{4},
\label{eq:mass}
\end{equation}
clamped at $\varepsilon{=}10^{-8}$ ($m_{i,j} = \max\{(1+3\alpha_i)/4,\varepsilon\}$)
so $\log m_{i,j}$ remains well-defined when $\alpha_i < -1/3$, while non-spatial
tokens keep unit mass ($m_{\mathrm{CLS}}{=}1$). Weighting each key in the
$4R$-slot training sequence by $m_j$ is equivalent to adding a 1D key bias
$\log m_j$ to the attention logits:
\begin{equation}
\begin{aligned}
\mathrm{Attn}(q,k,v;m)
&\;=\;
\frac{\sum_j m_j \exp\!\bigl(q^\top k_j/\sqrt{d_h}\bigr) v_j}
     {\sum_j m_j \exp\!\bigl(q^\top k_j/\sqrt{d_h}\bigr)}\\
&\;=\;
\frac{\sum_j \exp\!\bigl(q^\top k_j/\sqrt{d_h} + \log m_j\bigr) v_j}
     {\sum_j \exp\!\bigl(q^\top k_j/\sqrt{d_h} + \log m_j\bigr)},
\end{aligned}
\label{eq:attn}
\end{equation}
implemented as one extra coordinate in fused FlashAttention/SDPA kernels with
$O(N)$ memory. While ToMe~\cite{bolya2023tome} merges $s$ tokens at test time
and adds $+\log s$ during inference, \dvsp{} acts in the opposite direction: we
add $\log m(\alpha_i) \in [-\log 4, 0]$ to the $4$ unmerged virtual slots
\emph{only during training} so their combined weight tracks
$4m(\alpha_i) = 1+3\alpha_i \in [1,4]$, whereas test-time inference runs on
packed physical tokens with $m{\equiv}1$ ($\log m{=}0$).

\begin{proposition}[Training--Inference Endpoint Exactness]
\label{prop:endpoint}
Consider a Vision Transformer with virtual slots
(Eqs.~\eqref{eq:content}--\eqref{eq:mass}), mass-conserving attention
(Eq.~\eqref{eq:attn}), and aligned attention windows where a coarse parent and
its four slots attend to the same parent set. Across any number of layers:
(i) \textbf{Dense equivalence ($\alpha{\equiv}1$):} $m_{i,j}{=}1$ everywhere,
reproducing the pre-trained dense ViT up to floating-point round-off.
(ii) \textbf{Packed-inference equivalence ($\alpha \in \{0,1\}^R$):} Under
token-wise MLP/norm layers and a $\mathrm{CLS}$ or mass-weighted readout, the
$4R$-slot training network with mass $m(\alpha)$ produces the exact same
layer-by-layer features and task output as the test-time variable-length network
executing $N(\alpha)$ packed unit-mass coarse and fine tokens.
\end{proposition}

\noindent\emph{Proof sketch.} At $\alpha_i{=}0$, parent $i$'s four training
slots carry identical $(q_i, k_i, v_i)$ and mass $1/4$; in Eq.~\eqref{eq:attn}
their four key terms sum to $\exp(q^\top k_i/\sqrt{d_h}) v_i$, matching the
single unit-mass coarse token in the packed test sequence, while their identical
queries receive identical outputs that stay identical through token-wise norms,
MLPs, residuals, and unmerging $\mathcal{U}$. Induction over blocks gives~(ii),
and $\alpha{\equiv}1$ gives~(i) (Appendix~\ref{app:mass}). $\square$

\subsection{Dense-Initialized Training and Test-Time Projection}
\label{sec:training}

% Table I and Fig.~\ref{fig:cocoqual} are deliberately placed at this
% Sec.~III-D source position: double-column floats must be encountered while
% page~4 is typeset to stack together on page~5. Do not move them back into
% Sec.~IV without re-checking that co-location.
\begin{table*}[t]
\centering
\caption{\textbf{COCO val2017 detection and instance segmentation} with
EVA-02-B ($9{,}216$ dense visual tokens). Each budget block lists the same
five systems in the same order, so every \dvsp{} row is read against the
four rows directly above it. Parentheses give the mean fraction of kept
tokens. $\Delta$ is relative to the unadapted dense detector. DART-Grid's
$N=N_h N_w$ grid uses the exact counts 4{,}225/3{,}600/3{,}306, within
$0.4\%$ of each tier.}
\label{tab:coco}
\small
\setlength{\tabcolsep}{4pt}
\begin{tabular}{@{}lcccccc@{}}
\toprule
Method & Tokens & Box AP & $\Delta$Box & AP$_S$ & Mask AP & $\Delta$Mask \\
\midrule
Dense EVA-02-B & 9{,}216 (100\%) & 58.93 & -- & 42.09 & 50.69 & -- \\
\midrule
\multicolumn{7}{@{}l}{\emph{Budget $\approx$4{,}200 tokens}} \\
\addlinespace[2pt]
\hspace{1em}Dense, adapted at $1{,}040^2$ & 4{,}225 (45.8\%) & 57.07 & $-1.86$ & 37.89 & 48.52 & $-2.17$ \\
\hspace{1em}APT, entropy, per budget & $\sim$4{,}200 (45.6\%) & 57.33 & $-1.60$ & 39.89 & 49.40 & $-1.29$ \\
\hspace{1em}DART-Flow, fixed $N$ & 4{,}235 (46.0\%) & 55.81 & $-3.12$ & 36.90 & 46.99 & $-3.70$ \\
\hspace{1em}DART-Grid, fixed $N$ & 4{,}225 (45.8\%) & 55.94 & $-2.99$ & 35.85 & 47.12 & $-3.57$ \\
\hspace{1em}\dvsp{} ($\lambda{=}0.1$) & 4{,}235 (46.0\%) & \textbf{58.61} & \textbf{$-0.32$} & \textbf{41.90} & \textbf{50.41} & \textbf{$-0.28$} \\
\midrule
\multicolumn{7}{@{}l}{\emph{Budget $\approx$3{,}600 tokens}} \\
\addlinespace[2pt]
\hspace{1em}Dense, adapted at $960^2$ & 3{,}600 (39.1\%) & 56.19 & $-2.74$ & 36.65 & 47.72 & $-2.97$ \\
\hspace{1em}APT, entropy, per budget & $\sim$3{,}600 (39.1\%) & 56.65 & $-2.28$ & 39.19 & 48.56 & $-2.13$ \\
\hspace{1em}DART-Flow, fixed $N$ & 3{,}613 (39.2\%) & 49.95$^\ddagger$ & $-8.98$ & 26.17 & 41.50$^\ddagger$ & $-9.18$ \\
\hspace{1em}DART-Grid, fixed $N$ & 3{,}600 (39.1\%) & 55.19 & $-3.74$ & 35.54 & 46.54 & $-4.14$ \\
\hspace{1em}\dvsp{} ($\lambda{=}0.2$) & 3{,}613 (39.2\%) & \textbf{58.22} & \textbf{$-0.71$} & \textbf{41.63} & \textbf{49.90} & \textbf{$-0.78$} \\
\midrule
\multicolumn{7}{@{}l}{\emph{Budget $\approx$3{,}300 tokens}} \\
\addlinespace[2pt]
\hspace{1em}Dense, adapted at $928^2$ & 3{,}364 (36.5\%) & 55.93 & $-2.99$ & 36.22 & 47.56 & $-3.13$ \\
\hspace{1em}APT, entropy, per budget & $\sim$3{,}300 (35.8\%) & 56.09 & $-2.84$ & 38.26 & 47.98 & $-2.71$ \\
\hspace{1em}DART-Flow, fixed $N$ & 3{,}303 (35.8\%) & 54.57 & $-4.36$ & 34.41 & 45.99 & $-4.70$ \\
\hspace{1em}DART-Grid, fixed $N$ & 3{,}306 (35.9\%) & 54.91 & $-4.02$ & 34.19 & 46.25 & $-4.44$ \\
\hspace{1em}\dvsp{} ($\lambda{=}0.3$) & 3{,}303 (35.8\%) & \textbf{57.72} & \textbf{$-1.21$} & \textbf{41.00} & \textbf{49.44} & \textbf{$-1.25$} \\
\bottomrule
\end{tabular}

\vspace{2pt}
{\footnotesize $^\ddagger$DART-Flow degrades at this budget; best of several re-runs.}
\end{table*}

\begin{figure*}[t]
\centering
\includegraphics[width=\textwidth]{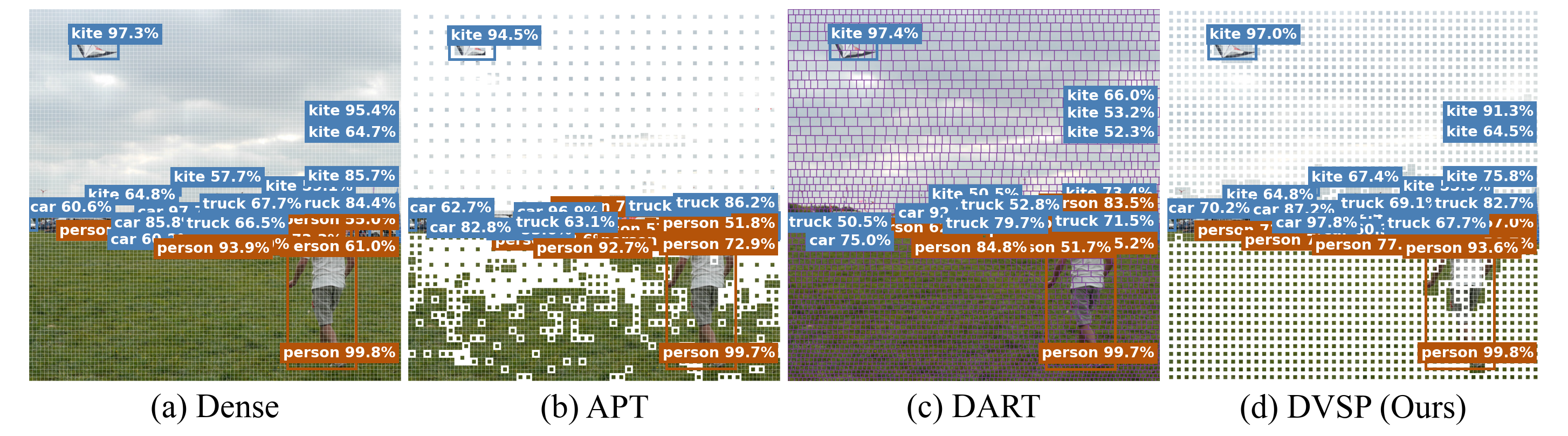}
\caption{\textbf{Token splits on a COCO scene.} (a) Dense, (b) APT,
(c) DART-Flow, (d) \dvsp{}, with $9{,}216$, $3{,}303$, $3{,}303$, and
$3{,}378$ tokens. Boxes with score $\ge0.5$. Of the $35$ annotated objects,
\dvsp{} matches $21$, as many as dense, against $14$ for APT and $13$ for
DART-Flow. More cases in Appendix~\ref{app:qualitative}.}
\label{fig:cocoqual}
\end{figure*}

\paragraph{Dense-initialized optimization}
By Proposition~\ref{prop:endpoint}(i), initializing the router's final linear
layer with zero weights and a $+1$ bias starts training at the exact pre-trained
dense model ($\alpha_i{=}1$). We jointly optimize the backbone and router under
\begin{equation}
\mathcal{L} \;=\; \mathcal{L}_{\mathrm{task}}(\alpha) \;+\;
\lambda(t)\,\frac{1}{R}\textstyle\sum_{i=1}^R |\alpha_i|,
\label{eq:loss}
\end{equation}
where $\lambda(t)$ warms up linearly from $0$ to $\lambda$. Leaving
$\alpha_i \in \mathbb{R}$ unconstrained avoids Sigmoid saturation near $0$ and
$1$, and naturally yields a bimodal distribution at the two binary endpoints
(Fig.~\ref{fig:coords}): on uninformative background where
$\partial \mathcal{L}_{\mathrm{task}}/\partial \alpha_i \approx 0$, the constant
$\ell_1$ subgradient ($+1$ for $\alpha_i > 0$) pushes $\alpha_i$ down until the
subdifferential sign flip traps it at $0$; on informative foreground, task
gradients pull $\alpha_i$ toward the pre-trained fine manifold at $1$, while
overshooting ($\alpha_i > 1$) exaggerates high-frequency residuals $(f_{i,j}-c_i)$
and raises $\mathcal{L}_{\mathrm{task}}$. Excursions past the endpoints do occur but
are shallow: across the trained epoch-50 fields, $6.4\%$/$7.6\%$ of
Tiny/Small coordinates leave $[0,1]$, yet fewer than $0.02\%$ fall below
$-0.1$ and under $0.7\%$ exceed $1$ (Fig.~\ref{fig:coords}); below $0$, the
$\varepsilon$ floor of Eq.~\eqref{eq:mass} keeps the extrapolated mass bounded.

\paragraph{From continuous training to discrete inference}
At test time, we discard the $4R$ virtual slots and $\log m$ bias, project each
parent's continuous coordinate onto its nearest binary vertex via
\begin{equation}
b_i^\star(x) \;=\; \mathbf{1}[\alpha_i(x) \ge \tfrac12],
\label{eq:projection}
\end{equation}
and execute the target variable-length pipeline of Sec.~\ref{sec:setup} on the
$N(b^\star)$ kept tokens. Because at least $96.5\%$ of trained coordinates already
sit near $\{0,1\}$ outside $[0.4, 0.6]$ (Fig.~\ref{fig:coords}) and
Proposition~\ref{prop:endpoint}(ii) guarantees zero gap at the binary vertices,
this continuous-to-discrete transition is near-lossless
(Sec.~\ref{sec:analysis}).

\section{Experiments}
\label{sec:experiments}

% Table II is placed here so the single-column [t] float is
% encountered while page 6 is built and lands at the page-6 column top.
% This splits the four tables across pages 5/6/7 instead of clustering
% Tables II-IV on page 7; do not move it back into Sec. IV-C without
% re-checking the page distribution.
\begin{table}[t]
\centering
\caption{\textbf{Frozen-task Cityscapes spatial selection.} Equal fine-parent
counts per crop. Binary rows execute all 2{,}304 slots at the projected binary vertices under
the mass rule of Eq.~\eqref{eq:mass}; by Proposition~\ref{prop:endpoint}(ii)
this equals packed native execution of the listed token counts layer by layer,
so counts measure the implied packed length, not executed slots. $\Delta$ is relative to dense.
Router checkpoint selected on validation.}
\label{tab:cityscapes}
\small
\setlength{\tabcolsep}{4pt}
\begin{tabular}{@{}lccrr@{}}
\toprule
Spatial score & Tokens (Red.) & mIoU & $\Delta$mIoU \\
\midrule
Dense reference & 2{,}304 (--) & 76.49 & -- \\
\midrule
APT entropy & 1{,}045 (54.6\%) & 72.38 & $-4.11$ \\
\textbf{\dvsp{}} & \textbf{1{,}045 (54.6\%)} & \textbf{74.77} & \textbf{$-1.72$} \\
\bottomrule
\end{tabular}
\end{table}

\subsection{Experimental Setup}
\label{sec:setup-exp}

Our experiments answer four questions in order.
\textbf{Q1:} does \dvsp{} work and run faster on the flagship dense-prediction
task? (Sec.~\ref{sec:coco}) \textbf{Q2:} does the gain survive without backbone
adaptation, on segmentation and classification? (Sec.~\ref{sec:frozen})
\textbf{Q3:} which component is responsible? (Sec.~\ref{sec:ablation})
\textbf{Q4:} why is continuous training safe for discrete deployment?
(Sec.~\ref{sec:analysis})

\textbf{COCO}~\cite{lin2014coco} is our end-to-end
accuracy--efficiency benchmark: an EVA-02-B~\cite{fang2023eva02} ViT-Adapter
detector at $1536{\times}1536$ ($9{,}216$ dense visual tokens), with wall-clock
speed measured under \texttt{torch.compile} + TensorRT including the router and
token packing. \textbf{Cityscapes}~\cite{cordts2016cityscapes} (frozen
Segmenter~\cite{strudel2021segmenter}) and \textbf{ImageNet-1k}~\cite{deng2009imagenet}
(frozen DeiT-Tiny/Small~\cite{touvron2021deit}, $196$ dense tokens at
$224{\times}224$) are controlled testbeds.

We compare against four adaptive baselines and one uniform control:
APT~\cite{choudhury2026apt}, the state-of-the-art entropy-based quadtree
detector, trained separately per budget; DART~\cite{yin2025dart} in both its
DART-Flow and DART-Grid variants, quantile-cut partitioners on our router's
exact frozen COCO trunk;
test-time ToMe~\cite{bolya2023tome} as an accuracy diagnostic;
and a uniform dense control adapted at the matched input resolution for the same
schedule. All COCO rows are scored with the same official postprocessing.
Appendix~\ref{app:baselines} gives adaptation details.

\subsection{Main Results: COCO Object Detection}
\label{sec:coco}

Table~\ref{tab:coco} reports detection and instance-segmentation accuracy on
COCO val2017 against every control at three token budgets; each budget block
lists the same five systems in the same order, so a \dvsp{} row is read against
the four rows directly above it. At $\lambda{=}0.3$, \dvsp{}
removes \textbf{$64.2\%$ of visual tokens} ($9{,}216 \to 3{,}303$) while
retaining \textbf{$97.9\%$ of dense Box AP} ($57.72$ vs.\ $58.93$) and
delivering a \textbf{$1.61\times$ end-to-end speedup}; at $\lambda{=}0.1$ it
stays within $0.32$ Box AP and $0.28$ Mask AP of dense using $46.0\%$ of the
tokens. Across every budget it more than halves APT's accuracy loss
($-0.32$ vs.\ $-1.60$ $\Delta$Box AP at $\sim$4{,}200 tokens), and at
$\sim$3{,}300 tokens it \textbf{beats APT at $\sim$4{,}200} on both Box and Mask
AP with $21\%$ fewer tokens. It likewise leads uniform lower-resolution dense
adaptation by $1.54$--$2.03$ Box AP at approximately matched counts.

The small-object column isolates the effect of staying on the regular lattice:
\dvsp{} retains 41.00 AP$_S$ against 38.26--39.89 for APT and 36.90 for
DART-Flow, while uniform $928^2$ adaptation keeps only 36.22 AP$_S$ at a
comparable budget. DART-Flow stays $2.8$--$3.1$ Box AP behind \dvsp{} where it
trains stably, and its $3{,}613$-token run drops to $49.95$ Box
AP$^{\ddagger}$. DART-Grid's shared-column grid stays $2.7$--$3.0$ Box AP
behind and keeps AP$_S$ at 34.19--35.85.

\paragraph{Qualitative comparison}
Fig.~\ref{fig:cocoqual} contrasts a rule-based allocator with a learned one. APT
places tokens by local pixel entropy; \dvsp{} places them by task gradient.
Dense spends $9{,}216$ tokens on this frame, and the three budget-matched panels
all spend about $3{,}300$, so those three differ in where tokens go, not how
many.
\dvsp{} is majority coarse here---$1{,}946$ of its $3{,}378$ tokens are
$32{\times}32$---yet it matches as many annotated objects as dense does, which
APT and DART-Flow do not. A budget-matched comparison on a textured wall isolates
the mechanism: APT covers $94\%$ of the wall above the objects
with fine $16{\times}16$ tokens, whereas \dvsp{} covers $93\%$ of it with coarse
$32{\times}32$ tokens, at $2{,}775$ tokens against APT's $4{,}950$---and all
four methods find the suitcase and the person there, so that scene separates
token placement from detection. The DART-Flow panel explains its
$2.8$--$3.1$ AP deficit as token geometry: its quantile cuts give $58$ rows
of unequal height
and off-lattice rectangles rather than the coherent regions a $3{\times}3$ head
consumes.

\begin{figure}[t]
\centering
\includegraphics[width=\columnwidth]{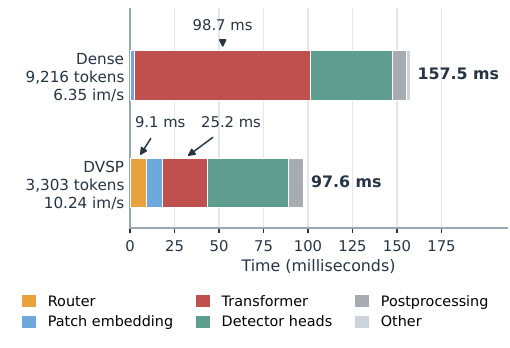}
\caption{\textbf{Measured EVA-02-B runtime breakdown.} COCO val2017, TITAN RTX,
FP16, batch one, same compile+TensorRT backend for both systems. Wall time
covers transfer, preprocessing, routing, patch construction, the detector,
and postprocessing, excluding data loading. Routing and the coarse-token
patch branch add $15.6$\,ms while the transformer drops $98.7\to25.2$\,ms,
so wall time falls $157.5\to97.6$\,ms ($6.35\to10.24$ images/s,
$1.61\times$).}
\label{fig:runtime}
\end{figure}

\paragraph{System runtime}
Fig.~\ref{fig:runtime} breaks down the measured end-to-end cost on one TITAN RTX
with the same compile+TensorRT backend for both systems. \dvsp{} raises
throughput from 6.35 to 10.24 images/s ($1.61\times$) at 3,303 tokens. Routing
($9.1$\,ms) and a coarse-token patch branch ($6.4$\,ms more than dense) add
15.6\,ms, but the transformer falls from 98.7 to 25.2\,ms, so routing
costs about a fifth of what the transformer saves and the net gain survives the
added work.

\emph{Takeaway (Q1).} On the flagship dense-prediction task \dvsp{} keeps
$97.9\%$ of dense Box AP at $64.2\%$ fewer tokens for a measured $1.61\times$
throughput gain, and beats every matched-budget control we ran; the small-object
column shows the retained accuracy is concentrated where the regular lattice
should matter most.

\subsection{Frozen-Backbone Controlled Studies}
\label{sec:frozen}

Detection adapts the backbone, so it cannot tell whether a gain comes from
better selection or from weights adapting around a poor selection. We therefore
pair two frozen-model studies that change exactly one variable each. On
Cityscapes the segmentation head is frozen and only the spatial partition
varies, isolating \emph{partition quality}; on ImageNet the classification
backbone is frozen and only the router is trained, isolating \emph{router
learning}. Both keep task weights fixed and compare at matched per-image
budgets, so neither can be explained by weights adapting around selection.

\begin{figure}[t]
\centering
\includegraphics[width=\columnwidth]{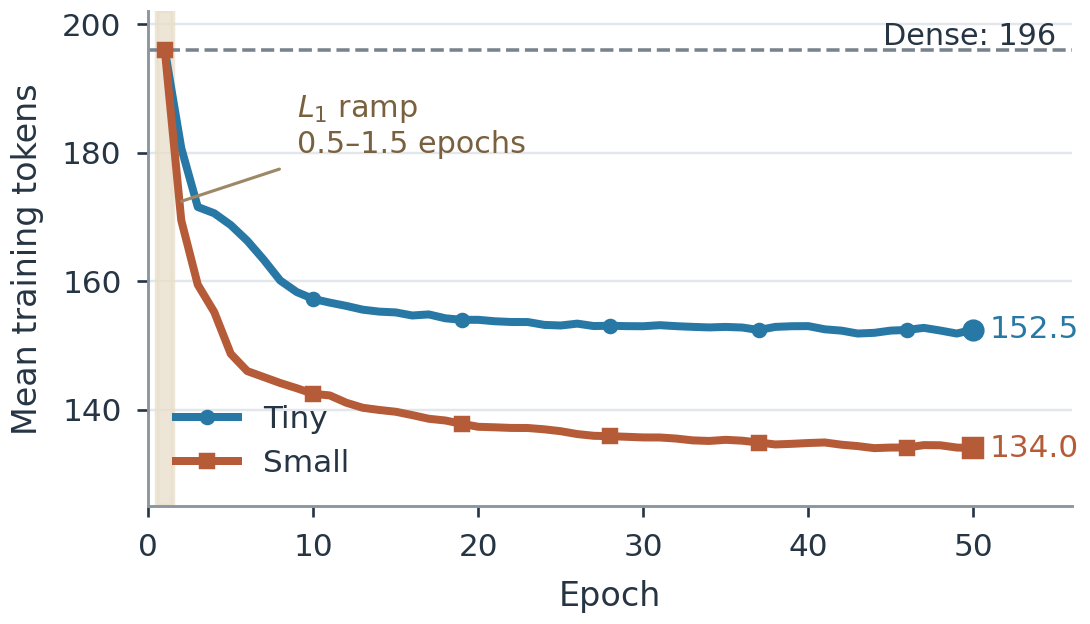}
\caption{\textbf{Token budget during router-only training on frozen DeiT
backbones (ImageNet-1k).} Training-set mean visual tokens; the $\ell_1$
penalty ramps over epochs $0.5$--$1.5$. Final full-validation Top-1 is
$75.36\%$ and $80.86\%$; executed validation lengths are $154.0$ and
$136.4$.}
\label{fig:curves}
\end{figure}

\paragraph{Spatial partition quality (Cityscapes)}
We evaluate a \emph{strictly frozen} Segmenter ViT-S/16 on $768{\times}768$
sliding windows over 500 validation images, giving the entropy baseline exactly
the same number of fine parent regions per crop as signed $\alpha_i\ge0.5$
\dvsp{}. Table~\ref{tab:cityscapes} shows \dvsp{} reaching \textbf{74.77\% mIoU}
against 72.38 for entropy: $+2.39$ mIoU at a 54.6\% implied token reduction.
Because the task model never updates, the difference is attributable to the
partition itself.
\paragraph{Router learning without backbone adaptation (ImageNet)}
To isolate the router, we freeze pre-trained DeiT-Tiny/Small and train
\emph{only} the router. From the all-fine state ($\alpha{\equiv}1$, $196$
tokens) the router drops $22\%$--$31\%$ of tokens during the one-epoch $\ell_1$
warmup, settling near $152.5$ (Tiny) and $134.0$ (Small) training-set mean
tokens (Fig.~\ref{fig:curves}). Final full-validation Top-1 under discrete
test-time execution is \textbf{75.36\%} and \textbf{80.86\%} against
$75.62\%$ and $81.46\%$ at the all-dense endpoint, so the reduction costs
$0.26$ and $0.60$ points.
At matched per-image counts ($154.0$ and $136.4$ validation tokens, CLS
excluded) \dvsp{} also exceeds entropy selection ($74.33\%$ and $79.83\%$)
by \textbf{1.03 Top-1 points on both backbones} (paired $95\%$ intervals
$[0.85,1.22]$ and $[0.84,1.22]$), and random selection at the same counts by
$2.74$ and $2.29$ points; the ordering also holds in cross-entropy
($0.9494$ vs.\ $0.9894$ and $0.6914$ vs.\ $0.7348$).

\emph{Takeaway (Q2).} With task weights frozen, both a segmentation head and a
classification backbone still favour \dvsp{}'s partition, so the gains are not
an artifact of weights adapting around router errors.

\subsection{Ablation Studies}
\label{sec:ablation}

\paragraph{Attention-mass conservation}
Sec.~\ref{sec:mass} argues that the affine mass rule $m(\alpha)=(1+3\alpha)/4$
is what lets the fixed-length training network agree with native execution.
Table~\ref{tab:ablation-mass} tests that claim on COCO: setting
$m\equiv1$---four unit-mass copies at $\alpha{=}0$, i.e.\ dropping the
$+\log m$ bias---costs \textbf{$1.59$ Box AP and $1.21$ Mask AP} at the same
$\lambda{=}0.3$ and a comparable token budget. Without mass conservation the
four collapsed slots exert four times the attention pull of a single native
coarse token, so the training objective stops matching the deployed network at
the coarse endpoint.

\begin{table}[t]
\centering
\small
\caption{\textbf{Attention-mass conservation on COCO val2017.} Same EVA-02-B
recipe and postprocessing; only the mass rule differs. Both rows execute in
native signed-threshold mode.}
\label{tab:ablation-mass}
\begin{tabular}{@{}lccc@{}}
\toprule
Mass rule & Tokens & Box AP & Mask AP \\
\midrule
Unit mass $m\equiv1$ & 3{,}268 & 56.13 & 48.23 \\
\textbf{Affine $m(\alpha)$ (ours)} & 3{,}303 & \textbf{57.72} & \textbf{49.44} \\
\bottomrule
\end{tabular}
\end{table}

\paragraph{$\ell_1$ weight to token budget}
The three operating points are three independently trained models, and the
mapping from $\lambda$ to budget is monotone and stable across the measured
range: $\lambda{=}\{0.3,0.2,0.1\}$ gives $3{,}303/3{,}613/4{,}235$ mean visual
tokens with $57.72/58.22/58.61$ Box AP and $49.44/49.90/50.41$ Mask AP. Varying this single scalar across otherwise identical runs therefore
sweeps the accuracy--budget curve, so one recipe covers all three COCO budgets
(Table~\ref{tab:coco}).

\emph{Takeaway (Q3).} Conserved attention mass is load-bearing rather than
decorative: removing it costs $1.59$ Box AP, more than any other single change
we measured, and a single $\lambda$ knob covers all three COCO budgets from
one recipe.

\begin{table}[!t]
\centering
\small
\caption{\textbf{Continuous field vs.\ its binary projection.} Each pair uses the
same router checkpoint and the same executed token count; accuracy is measured
with the field executed continuously ($\alpha$) versus projected
($b_i{=}\mathbf{1}[\alpha_i{\ge}0.5]$).}
\label{tab:projection}
\setlength{\tabcolsep}{3.5pt}
\begin{tabular}{@{}llrrr@{}}
\toprule
Task & Backbone & Continuous & Binary & $\Delta$ \\
\midrule
\multicolumn{5}{@{}l}{\emph{ImageNet-1k, top-1}} \\
\hspace{1em}Tiny, run 1 & DeiT-T & 75.06 & 75.33 & $+0.27$ \\
\hspace{1em}Tiny, run 2 & DeiT-T & 75.02 & 75.29 & $+0.27$ \\
\hspace{1em}Small, run 1 & DeiT-S & 80.74 & 80.86 & $+0.12$ \\
\midrule
\multicolumn{5}{@{}l}{\emph{Cityscapes, mIoU}} \\
\hspace{1em}resize, $\lambda{=}0.1$ & seg.-B & 74.23 & 74.89 & $+0.66$ \\
\hspace{1em}APT-embed, $\lambda{=}0.1$ & seg.-B & 74.28 & 74.93 & $+0.65$ \\
\hspace{1em}resize, $\lambda{=}0.3$ & seg.-B & 70.92 & 71.66 & $+0.74$ \\
\midrule
\multicolumn{5}{@{}l}{\emph{COCO, Box AP}} \\
\hspace{1em}$\lambda{=}0.1$ & EVA-02-B & 58.53 & 58.61 & $+0.07$ \\
\hspace{1em}$\lambda{=}0.2$ & EVA-02-B & 58.02 & 58.22 & $+0.20$ \\
\hspace{1em}$\lambda{=}0.3$ & EVA-02-B & 57.48 & 57.72 & $+0.24$ \\
\bottomrule
\end{tabular}
\end{table}

\subsection{Why Continuous Training Works for Discrete Testing}
\label{sec:analysis}

We verify the mechanism that makes unconstrained training safe at test
time (Table~\ref{tab:projection}).

\paragraph{Projection is near-lossless, and can help} Because most coordinates
already sit at an endpoint (Fig.~\ref{fig:coords}), snapping $\alpha$ to
$b_i = \mathbf{1}[\alpha_i \ge 0.5]$ leaves most regions untouched. Table~\ref{tab:projection}
compares the continuous field against its projection on nine checkpoints
spanning classification, segmentation, and detection, each at its own executed
token count: every change improves accuracy, by up to $0.74$ points. The per-image change still goes both ways.
$\Delta\mathrm{CE} = \mathrm{CE}(b) - \mathrm{CE}(\alpha)$ averages
$\overline{\Delta\mathrm{CE}} = -0.008$, with mean $95\%$ intervals within
$[-0.011,-0.0058]$ on all three training runs, so it is slightly negative rather
than merely near zero. The change is two-sided---$52\%$ of images improve and
$48\%$ worsen, with a median of $-7\times10^{-5}$---and only about $5\%$ of
images change their prediction at all, of which $758$ are corrected against
$625$ harmed on Tiny, with both other runs agreeing in direction.

Appendix~\ref{app:qualitative} gives full-page examples of the resulting
partitions for COCO, ImageNet-1k, and Cityscapes.

\section{Limitations}
\label{sec:limitations}

Our main detectors use a two-scale ($2{\times}2$) quadtree grid; while we
formulate a three-scale version ($32{\times}32$, $16{\times}16$, and $8{\times}8$
patches) with an exact dynamic-programming solver in the Appendix, deeper trees
remain future work. Second, while router runtime is fully included in our
$1.61\times$ speedup at $1536^2$ resolution, its fixed kernel launch cost takes
up a larger share of total time on small $224^2$ images. Finally, our ImageNet
experiments keep the ViT backbone frozen to cleanly measure single-region loss
changes; joint backbone fine-tuning for classification is left for future work.

Our COCO models are single runs, so the margins in Table~\ref{tab:coco} carry
no seed variance; the $\lambda$ sweep ($3{,}303$/$3{,}613$/$4{,}235$ tokens
with monoto\-nically increasing AP) indicates that the ranking is stable to
optimization noise. We have not run Sigmoid or initialization ablations of the
training path; the main claims are scoped to the two-scale,
unconstrained-$\alpha$ recipe we measured.

\section{Conclusion}
\label{sec:conclusion}

Prior methods for adaptive tokenization on regular grids have had to choose
between non-differentiable rules and noisy discrete masks. \emph{Differentiable
Virtual-Slot Partitioning} (\dvsp{}) resolves this trade-off by representing
coarse and fine quadtree states as the binary endpoints of a continuous,
fixed-length virtual-slot network. By controlling each parent's four virtual
slots---their token features and attention mass---with a single
continuous weight $\alpha_i$, \dvsp{} trains with exact gradients and deploys by
projecting onto the nearest binary vertex into a fast variable-length sequence.
Cutting $64.2\%$ of visual tokens on COCO for a $1.61\times$ wall-clock
speedup, \dvsp{} shows that regular grids and exact gradients work naturally
together.

\subsubsection*{Reproducibility Statement}
All experiments are implemented in the \texttt{dvsp} codebase submitted with
this paper, including model definitions, routers, mass-conserving Flash/SDPA and
Triton attention kernels, single-region importance evaluation scripts, and
testing pipelines. Every number in the main text comes directly from a verified
figure asset, log file, or experiment record in the code release. All experiments use public
datasets (COCO, ImageNet-1k, Cityscapes).

\bibliographystyle{IEEEtran}
\bibliography{refs}

% Appendices follow the references (arXiv preprint order, intended here).
% Only if this is later submitted to an IEEE venue as camera-ready, move the
% block below above \bibliographystyle.
\appendices

\section{Mass Conservation and Endpoint Exactness Details}
\label{app:mass}

\paragraph{Why naive slot duplication fails} In a standard variable-length ViT,
every token has a weight of $1$ in the attention softmax: a coarse parent gives
one token of mass $1$, while a fine parent gives four tokens of mass $1$ (total
mass $4$). If we represent a coarse parent inside a fixed-length $4R$ sequence
by placing four identical copies of $c_i$ with weight $1$, those four copies
together get $4\times$ as much attention probability as a single real coarse
token. Giving each coarse copy a mass of $1/4$ cancels
this $4\times$ duplication.

\paragraph{Uniqueness of the linear mass rule} Equation~\eqref{eq:mass} sets
the per-slot mass to $m(\alpha) = \frac{1+3\alpha}{4}$, which is the only linear
function $m(\alpha) = a + b\alpha$ that satisfies $m(0)=\frac{1}{4}$ and
$m(1)=1$. Under this rule, the total mass of a parent's four slots is
$4m(\alpha) = 1+3\alpha = (1-\alpha)\cdot 1 + \alpha \cdot 4$, which linearly
blends the total endpoint masses across $[0,1]$. While the $+\log m$ score bias
has the same math form as ToMe's proportional attention~\cite{bolya2023tome},
here it serves as a continuous, trainable weight over fixed virtual slots rather
than a fixed post-merge token count.

\paragraph{Proof of Proposition~\ref{prop:endpoint} (block-wise induction)}
Fix a compatible binary field $\alpha \in \{0,1\}^R$ and write $b_i = \alpha_i$
for its deployment switch. Let $\mathcal{N}_{\mathrm{slot}}$ be the $4R$-slot
network executing Eqs.~\eqref{eq:content}--\eqref{eq:attn} and
$\mathcal{N}_{\mathrm{pack}}$ the variable-length network of
Sec.~\ref{sec:setup} executing $N(\alpha)$ unit-mass tokens: parent $i$ emits
one coarse token $(c_i, p_i^c)$ if $b_i{=}0$ and its four fine tokens
$\{(f_{i,j}, p_{i,j}^f)\}_{j=1}^4$ if $b_i{=}1$. Two conditions are required.
\emph{(a) Duplicate preservation:} norms, MLPs, residuals, and the unmerging
$\mathcal{U}$ act independently per token position, so identical inputs stay
identical, and $\mathcal{U}$ copies a coarse output to all four child cells.
\emph{(b) Aligned attention windows:} in every windowed block each window is a
union of whole $2{\times}2$ parent blocks, and the window containing the four
slots of parent $i$ contains exactly the packed tokens whose windows contain
the corresponding packed coarse or fine tokens.

Call the slot state a \emph{grouped equivalent} of the packed state if, for
every packed token $t$, there is a disjoint set $S_t$ of slot positions holding
identical $(h, q, k, v)$ with $\sum_{s \in S_t} m_s = 1$. At the input this
holds: a coarse parent contributes four identical copies of $(c_i, p_i^c)$ with
$m{=}1/4$ each ($4 \times \tfrac14 = 1$), a fine parent its four tokens with
$m{=}1$ each, and positions agree by Eq.~\eqref{eq:position} at
$\alpha_i \in \{0,1\}$.

\emph{Attention preserves grouped equivalence.} Take any packed query $q_t$;
by (b), its copies in $S_t$ attend over exactly the same key set. Grouping the
softmax terms of Eq.~\eqref{eq:attn} by the sets $S_{t'}$, the identical
$(k_{t'}, v_{t'})$ within each set factor out and
$\sum_{s \in S_{t'}} m_s \exp(q^\top k_{t'}/\sqrt{d_h}) =
\exp(q^\top k_{t'}/\sqrt{d_h})$, so numerator and denominator collapse
term-by-term to those of the packed unit-mass token: every copy in $S_t$
receives exactly the packed output $o_t$. Windowed and global blocks share the
same argument within their respective token sets.

\emph{Token-wise layers preserve grouped equivalence.} Norms, MLPs, and
residuals act per position and keep the copies in $S_t$ identical to the packed
feature $h_t$; $\mathcal{U}$ then places $h_t$ (or, at coarse parents, the
shared coarse feature) at each child cell. The readout --- a $\mathrm{CLS}$
token attending globally, or mass-weighted pooling --- sums identically over
grouped copies. By induction over blocks, $\mathcal{N}_{\mathrm{slot}}$ and
$\mathcal{N}_{\mathrm{pack}}$ produce identical layer-by-layer features and
task outputs, proving~(ii). Claim~(i) is the special case $\alpha \equiv 1$:
all masses equal $1$ and the slots are the native fine tokens, so both networks
reduce to the unmodified dense ViT. The $\varepsilon$ floor of
Eq.~\eqref{eq:mass} is inactive on $\{0,1\}^R$, so the clamp never enters the
identity. $\square$

\paragraph{How the aligned-window condition is met in our detector}
In our EVA-02-B detector, windowed blocks use windows of an even number of fine
cells, so window boundaries fall on parent boundaries and no $2{\times}2$
parent is ever split across windows. The packed native path then runs the same
blocks on the coarse grid with windows of half the linear size, which makes a
coarse token's window contain exactly the parents whose four slots share one
fine window; remaining blocks are global, where the condition holds trivially.

\section{Exact-Budget Projection for Three-Scale Hierarchies}
\label{app:threescale}

While our main detection models use simple per-parent thresholding
($\alpha_i \ge 0.5$), \dvsp{} also extends to three-scale quadtrees (with
$32{\times}32$ root tokens, $16{\times}16$ middle tokens, and $8{\times}8$ fine
tokens inside each $4{\times}4$ root block). When an exact total token count is
needed across a three-scale tree, we cast the discrete choice as a
multiple-choice knapsack problem: each root region picks among valid subtree
splits costing $\{1,4,7,10,13,16\}$ tokens, scored by the sum of its continuous
weights. A fast dynamic-programming (DP) solver across roots finds the best
global split for the exact target token count. We provide both a standard nested
tree solver and an overlap solver that shares coarse tokens across sibling
subtrees (81 patterns), along with single-region flip tests for evaluation.

\section{Additional Qualitative Results}
\label{app:qualitative}

\begin{figure*}[t]
\centering
\includegraphics[width=\textwidth]{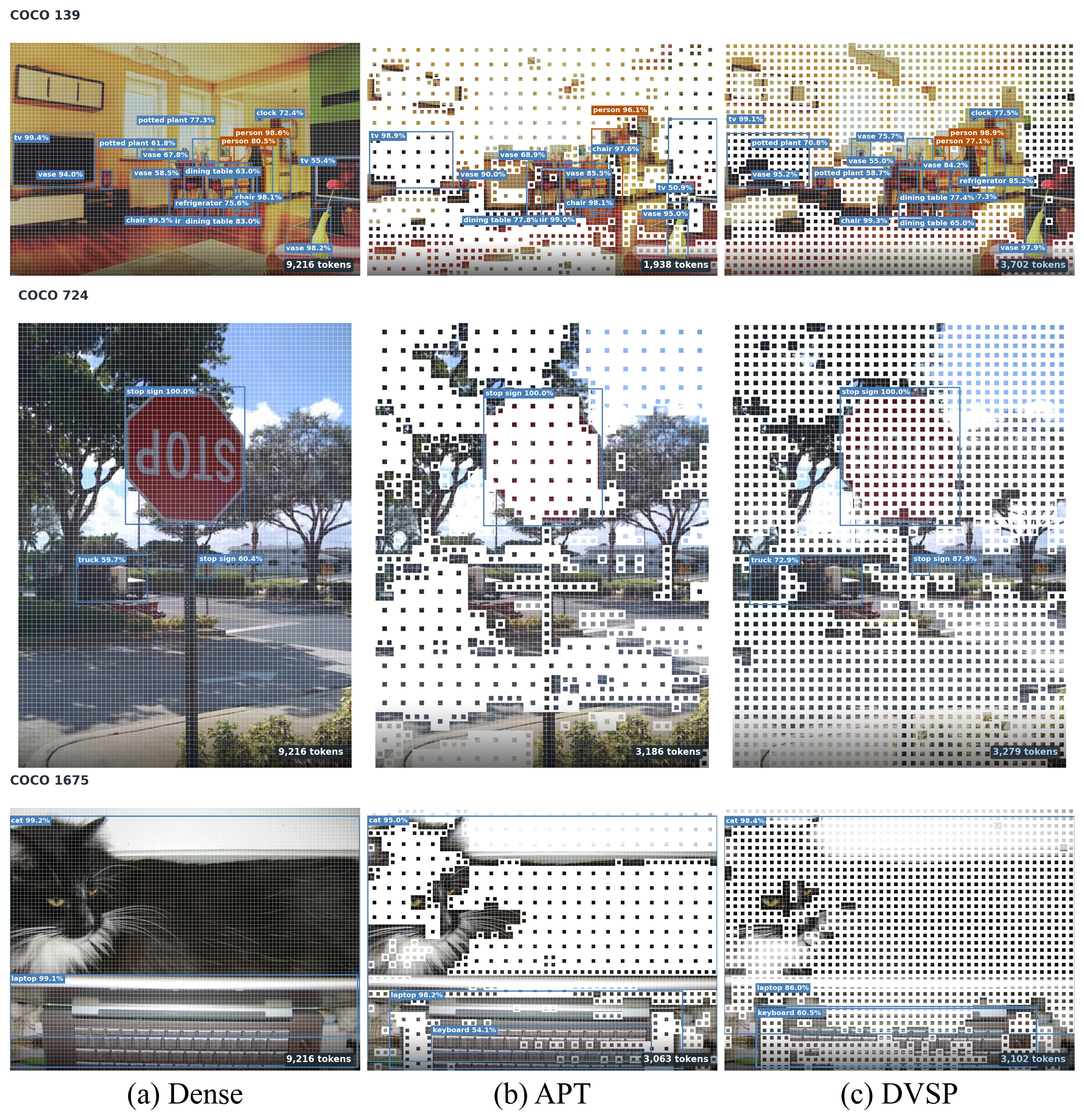}
\caption{\textbf{COCO object detection examples.} Each row is one val2017
scene; columns show the Dense detector, APT, and \dvsp{} with their recorded
partitions (white lines, $32{\times}32$ coarse regions rendered as centered
mini patches) and all saved score $\ge0.5$ predictions. Token plates give each
model's executed count on the full $1536^2$ canvas. Best viewed zoomed in.}
\label{fig:cocoexamples}
\end{figure*}

\begin{figure*}[t]
\centering
\includegraphics[width=\textwidth,height=0.73\textheight,keepaspectratio]{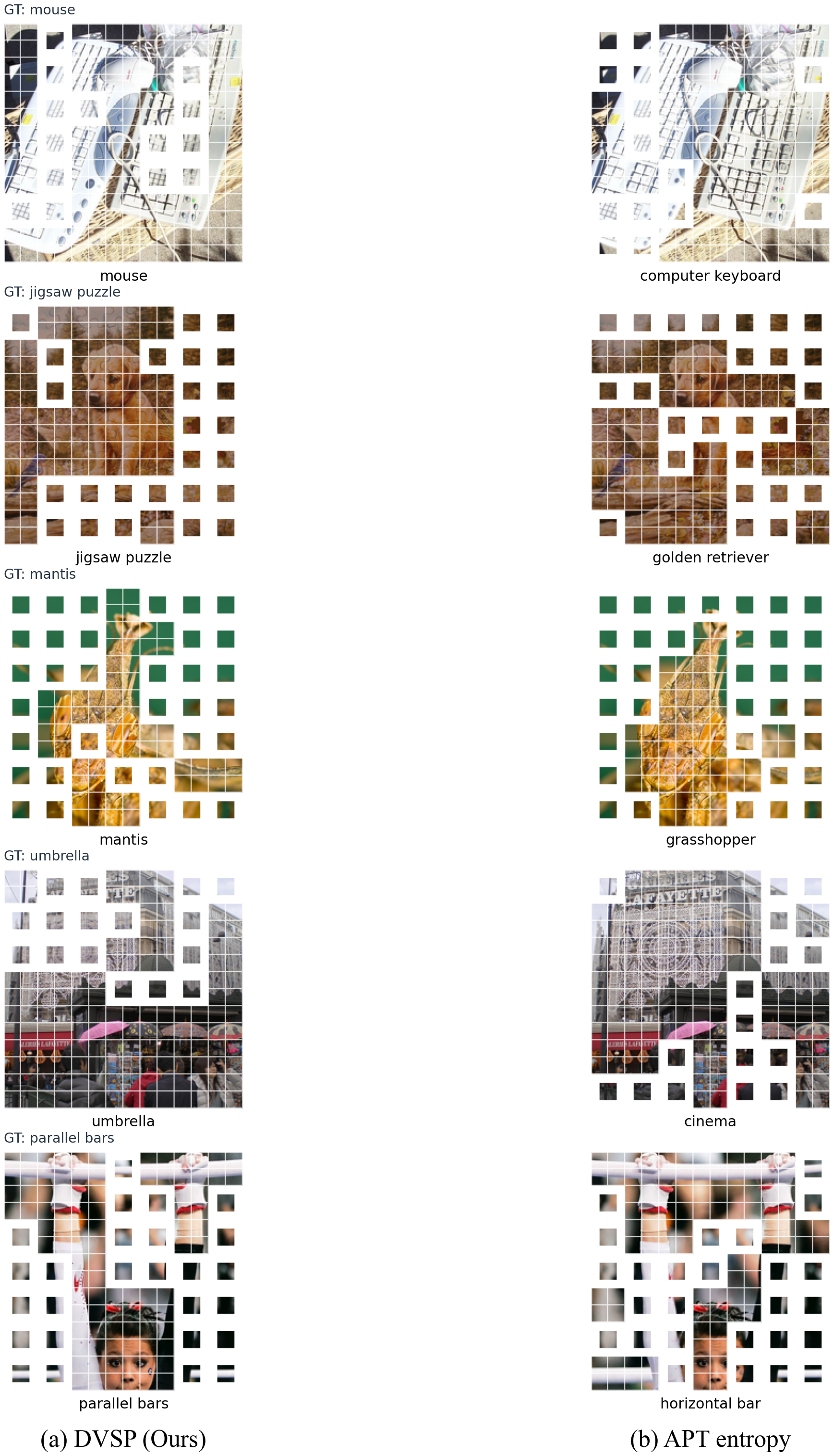}
\caption{\textbf{ImageNet-1k classification examples.} Five val images from the
frozen-backbone router study of Sec.~\ref{sec:frozen}; in every row \dvsp{} is
correct while the entropy control is wrong. (a)~\dvsp{} partition;
(b)~APT entropy partition. Class names under the panels give each method's
Top-1 prediction, and the ground truth tags each row. \dvsp{} spends its tokens
on the object while the entropy control spends them on background texture.
Best viewed zoomed in.}
\label{fig:clsexamples}
\end{figure*}

\begin{figure*}[t]
\centering
\includegraphics[width=\textwidth,height=0.55\textheight,keepaspectratio]{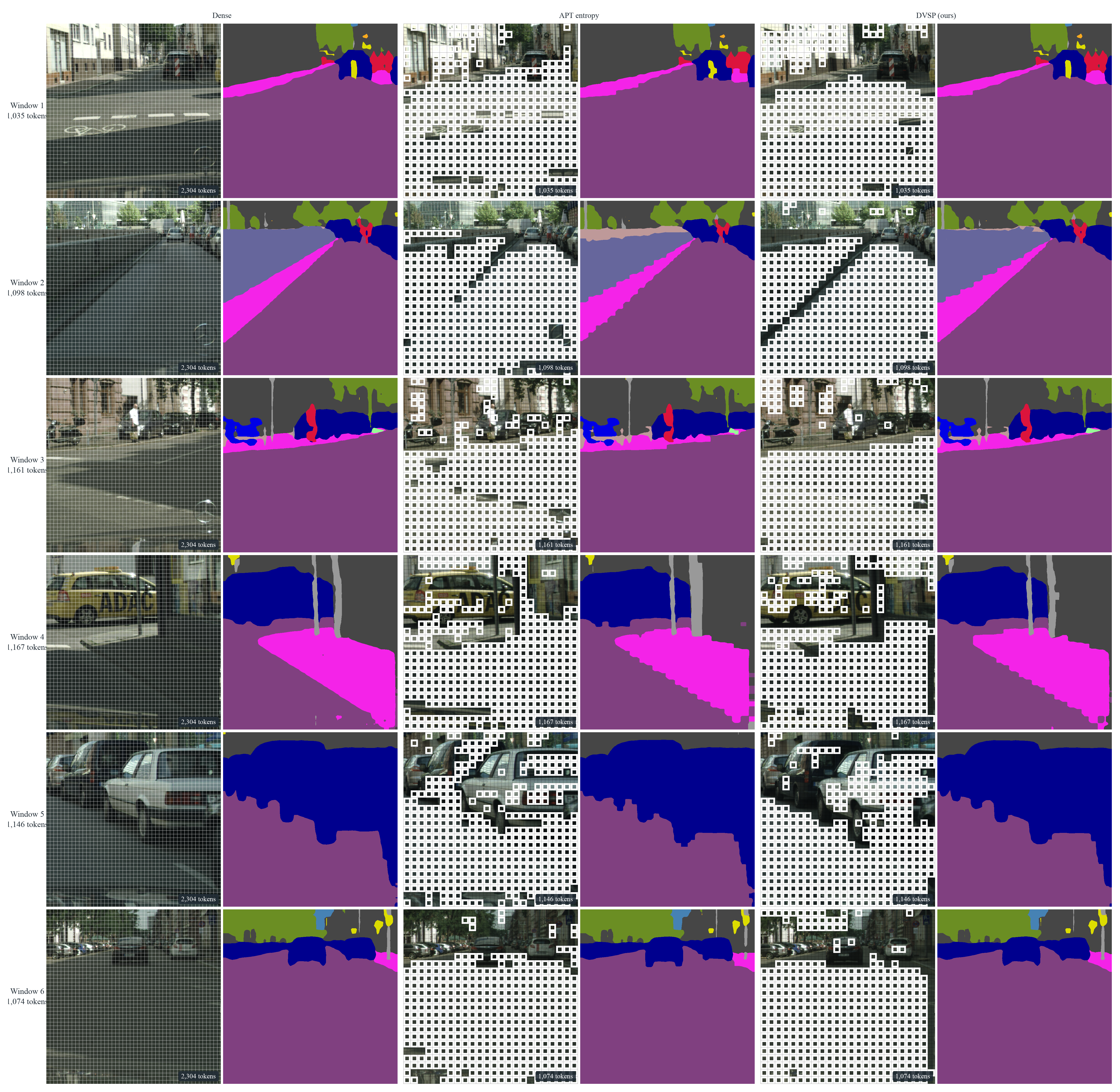}

\caption{\textbf{Cityscapes semantic segmentation examples.} Six fixed
$768{\times}768$ val windows (rows) under the frozen Segmenter of
Sec.~\ref{sec:frozen}: two non-overlapping windows from each of three val
images. Each window compares three methods at the same fine-parent budget:
Dense (uniform $16{\times}16$ grid, $2{,}304$ tokens), APT entropy (the
Table~\ref{tab:cityscapes} baseline), and \dvsp{}
($1{,}035$/$1{,}098$/$1{,}161$/$1{,}167$/$1{,}146$/$1{,}074$ tokens). In each of the three columns the left panel is the input with that
method's executed patch assignment and the right panel the resulting
segmentation. Entropy places fine patches on textured facades and foliage,
while \dvsp{} keeps them along object edges and small pedestrians; flat road
and sky become $32{\times}32$ tokens in both adaptive rows. The \dvsp{}
partitions use the recorded $|\alpha|\!\ge\!0.1$ projection of this
frozen-scorer study.}
\label{fig:segexamples}
\end{figure*}

One page of examples per deployment dataset: COCO object detection
(Fig.~\ref{fig:cocoexamples}), ImageNet-1k classification
(Fig.~\ref{fig:clsexamples}), and Cityscapes semantic segmentation
(Fig.~\ref{fig:segexamples}). Best viewed zoomed in.

\section{Implementation and Training Details}
\label{app:details}

\paragraph{Building coarse tokens} Following APT~\cite{choudhury2026apt},
scale-32 coarse tokens $c_i$ are built by running the ViT patch embedding on a
downsampled ($0.5\times$) copy of the input image, plus a learnable residual
branch made of a $2{\times}2$ strided conv over the fine token grid followed by a
\emph{zero-initialized} linear layer. At Step 0, $c_i$ is simply the
downsampled patch embedding.

\paragraph{Router architectures}
(i) \emph{Detection (Main):} We extract frozen stride-16 features from
MobileNetV3-Large, combine spatial and semantic stages with $1{\times}1$
convs and a small fusion CNN, and resize the output to the $R$-parent grid.
Router weights are computed in FP32 on the padded image.
(ii) \emph{Detection (DART-trunk version):} We freeze the official DART scorer
backbone, reset its last MLP layer to zero weights and $+1$ bias, and bilinearly
resize its output to the parent grid.
(iii) \emph{Classification analysis:} We attach a small MLP router (or a 5-block
GroupNorm+GELU conv router) to frozen DeiT features.

\paragraph{Training schedules} On ImageNet-1k classification, routers are
trained for 50 epochs with the $\ell_1$ weight $\lambda(t)$ warmed up linearly
from $0$ between epochs $0.5$ and $1.5$. On COCO object detection
(Sec.~\ref{sec:coco}), we train separate models with fixed
$\lambda \in \{0.1, 0.2, 0.3\}$.
\paragraph{Analysis tools} The diagnostics in Sec.~\ref{sec:analysis} come from
logging hooks built into the training loop: 16-bin histograms of $\alpha$ on
$[-1,1]$, continuous-vs-discrete $\Delta\mathrm{CE}$ checks, single-region flip
tests, and gradient similarity checks between $\mathcal{L}_{\mathrm{task}}$ and
the $\ell_1$ penalty.

\paragraph{Ablation infrastructure}
The code release also includes implementations for:
three-scale router options (unconstrained product, bounded product, group
softmax, and slot-wise softmax); Bernoulli REINFORCE routers; Gauss--Legendre
path-integral importance probes; rank-based penalty weighting; frozen patch/CLS
feature distillation; LoRA backbone adapters; dual-projection patch embeddings;
and ShuffleNet-trunk routers. Selected ablation tables remain in the project
experiment records.
\section{Baseline Adaptations}
\label{app:baselines}

\textbf{APT}~\cite{choudhury2026apt} uses its official grayscale histogram
entropy rule (512 bins) to pick patch scales; to make it a strong baseline on
COCO, we train a separate APT detector for each token budget using its official
multi-resolution sampling schedule ($\{0, 0.25, 0.5, 0.75, 1.0\}$).
\textbf{ToMe}~\cite{bolya2023tome} is adapted to windowed EVA-02 detection for
test-time evaluation: bipartite matching and weighted merging are applied
between attention and MLP layers inside each window (and in global blocks),
keeping target RoPE positions and unmerging tokens back to the full grid before
the detection head. The adapted variant reaches 56.74 Box AP at a 3,303
final-token budget, but its per-layer attention input still averages about 6,496
tokens and its measured latency is 263.6 ms/image versus 243.6 ms for dense.
\textbf{DART}~\cite{yin2025dart} is evaluated in both variants on our router's
exact frozen COCO trunk. DART-Flow (the main method) cuts one global quantile
sequence, so columns differ across rows; DART-Grid (official
\texttt{DynamicAdaptiveImageReshaper}) quantizes the row and column marginals
independently into an $N_h{\times}N_w$ grid with one shared column set. We adapt
its 2D-separable geometry to the $96{\times}96$ EVA grid; both variants use
the original dense weights and the same 3000-update BF16 recipe.

\end{document}